\documentclass[sigconf]{acmart}

\renewcommand\footnotetextcopyrightpermission[1]{} 
\usepackage{tikz}
\usepackage{amssymb}
\usepackage{siunitx}
\usepackage{booktabs}
\usepackage{graphicx}
\usepackage{tabularx}
\usepackage{url}
\usepackage{caption}
\usepackage{afterpage}

\newcommand{\eg}{\textit{e.g.}}
\newcommand{\ie}{\textit{i.e.}}
\newcommand{\etc}{\textit{etc}}

\makeatletter
\def\@copyrightspace{\relax}
\makeatother
\begin{document}
\title{Exploration and Exploitation in Symbolic Regression using Quality-Diversity and Evolutionary Strategies Algorithms}

\author{J.-P. Bruneton}
\authornote{jpbruneton@gmail.com (corresponding author)}
\affiliation{%
  \institution{LIED, Universit\'e de Paris \\ CNRS/UMR 8236}
  \city{Paris} 
  \state{France} 
  \postcode{F-75013}
}

\author{L. Cazenille}
\authornote{leo.cazenille@gmail.com}
\affiliation{%
  \institution{Ochanomizu University}
  \city{Tokyo} 
  \state{Japan} 
}

\author{A. Douin}
\authornote{adele.douin@gmail.com}
\affiliation{%
  \institution{LIED, Universit\'e de Paris \\ CNRS/UMR 8236}
  \city{Paris} 
  \state{France} 
  \postcode{F-75013}
}

\author{V. Reverdy}
\authornote{vince.rev@gmail.com}
\affiliation{%
  \institution{University of Illinois at Urbana-Champaign}
  \streetaddress{Astronomy Department, 1002 West Green Street}
  \state{Illinois}
  \country{USA}
}

\begin{abstract}
By combining Genetic Programming, MAP-Elites and Covariance Matrix Adaptation Evolution Strategy, we demonstrate very high success rates in Symbolic Regression problems. MAP-Elites is used to improve exploration while preserving diversity and avoiding premature convergence and bloat. Then, a Covariance Matrix Adaptation-Evolution Strategy is used to evaluate free scalars through a non-gradient-based black-box optimizer. Although this evaluation approach is not computationally scalable to high dimensional problems, our algorithm is able to find exactly most of the $31$ targets extracted from the literature on which we evaluate it.
\end{abstract}

\maketitle

\section{Introduction}
\label{sec:introduction}
Symbolic Regression (SR) aims at building mathematical models of numerical, and possibly experimental, data. Given data of the form $(y_i, \vec{x}_i)$, the goal is to discover automatically the analytical relationship between $y$ and $x$, as a function of a mathematical language that usually includes basic operators like $(+,-,\times,/)$, possibly also non-algebraic functions such as sine, exp, ..., some free scalars (pure numbers), and the variables $\vec{x}$. 

If such an analytic relation exists by construction, \ie{} when we feed the program with data $y_i = f_{\textrm{target}}(x_i)$, then the goal is to write candidate equations $\tilde{f}(\textrm{vocabulary})$ until one hits the target $\tilde{f} = f_{\textrm{target}}$. If not, the goal is to find a good approximation of the target function on the provided data (the \emph{training set}) with good generalization properties, meaning that we want the discovered functions to behave well on previously unseen data, known as the \emph{test set} or \emph{validation set}.

Besides being a difficult machine learning problem that is interesting on its own, SR can also be used to provide accurate models of physical systems that are too complex to be theoretically modelled. Any complex phenomena emerging from the underlying dynamics of a large number of degrees of freedom typically fall in this category. This happens in particular in the fields of meteorology, climatology, material properties, heat transfers, astrophysics, economy, financial data, complex systems, etc. Notice that the vocabulary can include derivatives with respect to the variables so that dynamics can also be discovered provided the data has some temporal component. Even in the case where the outcome of the program is not a perfect fit, finding accurate solutions may guide the researcher towards a better understanding of the system's underlying physics. In this regard, the interpretability of the fittest candidate equations is important. We shall comment on this later on.

SR has been studied quite extensively along these lines. For instance, references~\cite{Bongard9943, Schmidt81} try to recover physical laws and invariants of some mechanical systems,~\cite{quade2016prediction} focuses on real-world complex systems data sets, while, related to our concern, ~\cite{2018arXiv181010525W} aims at building an "automated physicist".

One of the main approach to SR is Genetic Programming (GP) which is based on a computer simulated Darwinian evolution\footnote{Notice that GP can of course be used to solve other problems than SR.}, see, \eg{}, textbooks \cite{koza1992genetic, poli2008field}. In this field, the vocabulary is known as the \emph{primitive set}, while the \emph{individuals} built from correspond to candidate equations. Some number of individuals are created initially, and then selected according to their \emph{fitness} (\ie{} some metric) with regards to the problem at hand, and then evolved by genetic operations, namely mutations of the equation, or crossovers between two equations. This scheme then iterates until the target is found or some computational budget is reached.  

Although GP has proven very successful for finding highly fit individuals in large search spaces, it has some long standing issues regarding in particular the recurrent loss of diversity during evolution (see Section \ref{sec:background:tree}). Recently however, a new paradigm emerged of evolutionary algorithms that aims at exploring both quality and diversity of individuals. These algorithms are not looking for the fittest individual only, but rather look for a set of high performing ones given their behavior with respect to hand-designed features (see Section \ref{sec:background:map}). This so-called MAP-Elites algorithm \cite{DBLP:journals/corr/MouretC15} has been used as an improvement of GP in algebraic problems \cite{dolson2019exploring}, path-finding \cite{pugh2016quality, 2018arXiv180702397G}, design discovery \cite{gaier2017data}, robotics \cite{duarte2018evolution}, and is available as a \textsc{Python} library \cite{qdpy}. As far as we are aware, it has not been applied to SR yet. Although later improvements have been proposed to the algorithm, \eg{} \cite{pugh2016searching, cully2018quality}, we shall restrict ourselves here to its original version as published in \cite{DBLP:journals/corr/MouretC15}.

In this paper, we will apply this enhanced exploration algorithm to the problem of Symbolic Regression. However, we are not only concerned here with maintaining diversity, but also with a better exploitation of the results. One striking issue concern the way free scalars in SR are handled. In most of the published literature, free scalars that appear while building an individual can either be picked up randomly from a given integer set, for instance $\{-2, -1, 0, 1, 2\}$, or be randomly chosen floating-point numbers in a predefined and fixed interval -- the so-called "ephemeral constants" \cite{Davidson:2003}. 

This, we believe, is not quite satisfactory. If we limit ourselves to integer scalars only, given that the equation has a maximal size, we cannot build all real-valued scalars this way. On the other hand, using ephemeral constants requires by construction many iterations of the evolution scheme before finding a value that is accurate enough.

Instead, we will write candidate equations with not yet assigned free scalars under forms such as $f = A_1 \times \exp(A_2 \times x)$ and then find the best scalars $A_i$. However, achieving this cannot rely on common gradient-based techniques since they would often converge to a local optimum and miss the global one\footnote{Still we note that \cite{kommenda2013effects, quade2016prediction} try to fit numerical constants with gradient descent and show that it is already an improvement over the use of ephemeral constants. On the other hand, \cite{cerny2008using} uses instead another non-gradient based method, but limits itself to very simple targets only (at most order three polynomials).}. Instead, we use another evolution algorithm, namely a Covariance Matrix Adaptation-Evolution Strategy algorithm~\cite{hansen2003reducing} (CMA-ES) in order to look for the best fit for the free scalars. Details of the method can be found in Section \ref{sec:background}. Because CMA-ES is computationally heavy, it could not be reasonably applied to very long equations (say of more than 60 elements) with many scalars. However, it represents a substantial improvement that is worth considering for mid-sized targets. 

To summarize, our model relies on an improved exploration of the fitness landscape via the evolutionary Quality-Diversity algorithm, and then fits the best scalars with another evolutionary algorithm (CMA-ES). This last technique is specific to SR and shall not apply to other types of problems that GP usually deals with. These two methods combined result in a very high success rate on many targets found in the literature. Moreover, even when the algorithm fails to find the exact target, it returns very accurate fits thanks to the CMA-ES method (although generalization may be poor in this case). Section \ref{sec:background} provides some background to both plain GP and its limitations and to the MAP-Elites algorithm. It finally gives a quick overview of how CMA-ES method works. Section \ref{sec:model} describes the entire model by putting together all these pieces, and Section \ref{sec:results} shows experimental results. 

\section{Background}
\label{sec:background}
\subsection{Tree-based GP for SR}
\label{sec:background:tree}
\begin{figure}[ht]
	\begin{centering}
		\begin{tikzpicture}[level/.style={sibling distance = 2cm/#1,
  level distance = 1cm}] 
		\tikzstyle{hollow node}=[circle, draw, inner sep = 4pt, align = center, fill = black!5]
		\node(0)[hollow node]{$\times$}
			
			child{node[hollow node]{$+$} child{node[hollow node]{$x$}} child{node[hollow node]{$y$}} }
			child{node[hollow node]{$x$}};
		\end{tikzpicture}
		\caption{Tree representation of $\left(x+y\right) \times x$ or $x \, y \, + \, x\, \times \, \varnothing$.\label{fig:tree}}
	\end{centering}	
\end{figure}
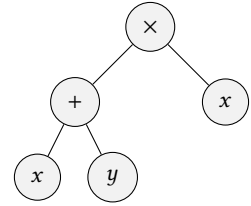

Before going to the GP algorithm and its improvements, let us first quickly outline how SR is usually implemented in so-called \emph{tree-based GP}. Mathematical expressions are created and modified either as strings of symbols, usually in infix or postfix (reverse Polish) notation, as Abstract Syntax Trees (ASTs), or as a combination of the two depending on which representation fits best each section of the SR algorithm. ASTs are especially convenient to run genetic operators such as crossovers and point-wise mutations. The SR algorithm is given a \emph{primitive set} of symbols, including $\varnothing$ that serves as a \emph{halt} symbol to terminate the expression. For example, in the postfix notation that we use, $\left(x+y\right) \times x$ is represented as $x \, y \, + \, x\, \times \, \varnothing$ and corresponds to the tree given in Fig. \ref{fig:tree}.
Basic mathematical rules can easily be encoded in ASTs. While the literature usually imposes a limit on the tree depth, we will instead impose a hard limit on the length $L$ of the mathematical expressions produced by the algorithm. This corresponds to a parsimony measure~\cite{koza1992genetic,iba1994genetic} of the generated equations. This choice was motivated by the fact that parsimony is taken directly into account into our methodology, see next subsection.

\begin{figure}[ht]
	\begin{centering}
		\includegraphics[width=\linewidth]{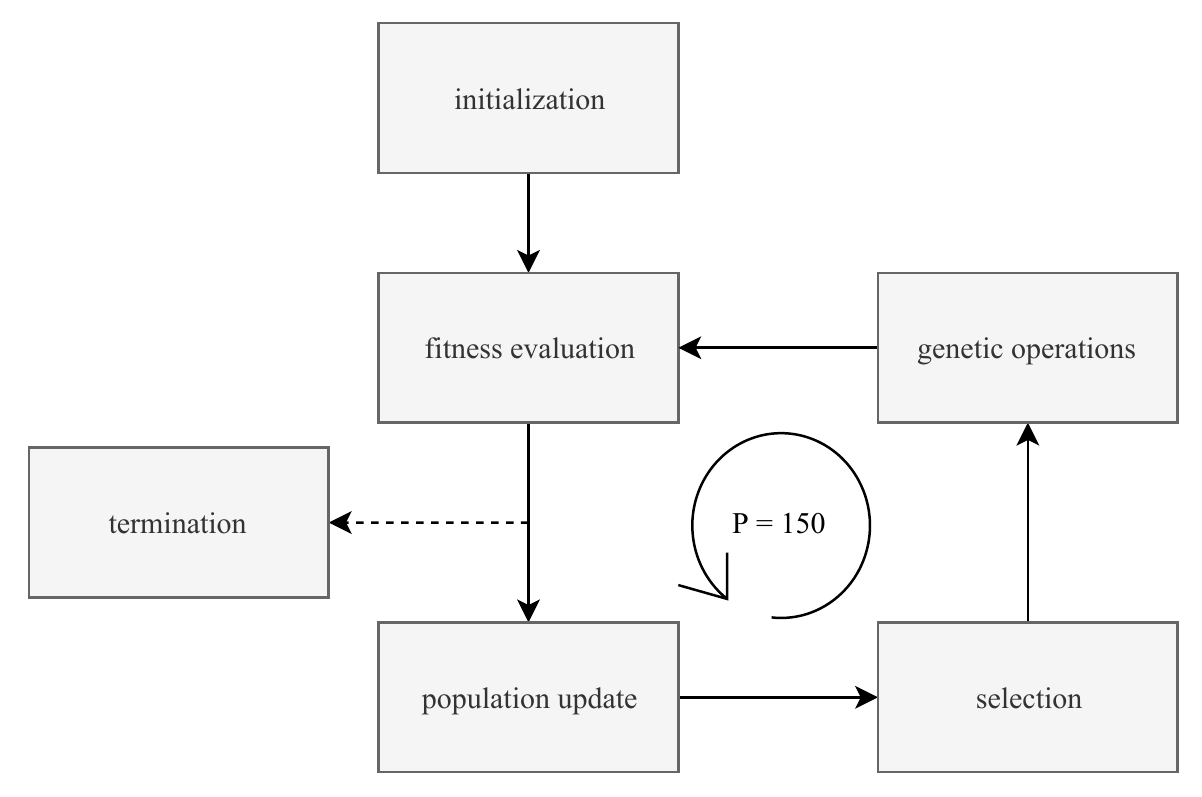}
        \caption{High-level view of Genetic Programming iterations.\label{fig:plaingp}}
	\end{centering}
\end{figure}

GP algorithms require the following ingredients, whose relationship are summarized in Fig. (\ref{fig:plaingp}): a population of individuals and its initialization, fitness evaluation, selection, genetic mutations, population update, and meta-parameters. A very crude view of GP is the following. After the creation of an initial population of $N$ random equations (see \cite{koza1992genetic} for variations of initialization techniques), individuals are evaluated with respect to the target on the \emph{training set}; then some are selected either randomly or in relation to their fitness. These individuals go through genetic operations, basically mutations and crossovers; these new individuals are evaluated, and the population is updated to keep only the $N$ best equations. The algorithm then iterates over this scheme.
Termination occurs when one individual is accurate enough with respect to the \emph{validation set}, see Section \ref{sec:model:termination} for details, or when the computational budget is exhausted.

Many recent papers focus on improving one or more of these steps. For instance, one may not want to only target the best fit individual, but other features as well. This has led to \emph{multi-objective optimization} and \emph{Pareto-front exploitation} where fitness evaluation considers several objectives at the same time, see \eg{} \cite{6791888, laumanns2002archiving, smits2005pareto}. While the algorithm runs, a single population of individuals is kept in memory. However, it might be useful to decompose hard problems into smaller, easier sub-problems. Therefore this scheme can also be tweaked to incorporate problem decomposition, by keeping small blocks of expressions that have proven useful during the training, see, \eg{}, \cite{koza1994genetic, arnaldo2014multiple, astarabadi2018decomposition}. As we shall see, MAP-Elites includes a sort of problem decomposition when remembering the small but fit individuals.

GP still has long standing issues, however. One of them is the bloat that happens when no hard limits are set on the length of expressions; then, crossovers tend to create longer individuals while their fitness no longer improves. Many techniques have been designed to counter bloat, in particular setting hard limits, or setting a "soft limit" by disadvantaging long genomes (\ie{} individuals), see, \eg{}, \cite{Bloat}, or even variations of this \cite{poli2003simple}. As we shall see, the MAP-Elites algorithm can automatically counter this problem.

Another issue is the premature convergence or diversity loss during evolution. This may be prevented by increasing population size, modifying genetic operations, and/or selection mechanism. As a first guide to the improvement of genetic operations besides the basic point-wise mutation and crossovers, we refer the reader to \cite{poli2008field} and references therein.

\subsection{MAP-Elites}
\label{sec:background:map}
MAP-Elites algorithm belongs to the class of Quality-Diversity algorithms \cite{pugh2016quality, cully2018quality} that also includes, for example, Novelty Search with Local Competition (NS-LC) \cite{lehman2011evolving, lehman2013effective}. The algorithm is grid-based and stores the best-fit individuals in a grid of elites; the grid being $N$-dimensional with $N$ features chosen by the user.
As far as equations are concerned, quite natural features one may think of includes the length of the equation (or other metrics of its complexity), the number of free scalars, the number of nested non-algebraic functions such as $\sin(\sin(...))$, the number of trigonometric functions, and the order of non-linearity of \cite{vladislavleva2009order}.

\begin{figure}[ht]
	\begin{centering}
		\includegraphics[width=\linewidth]{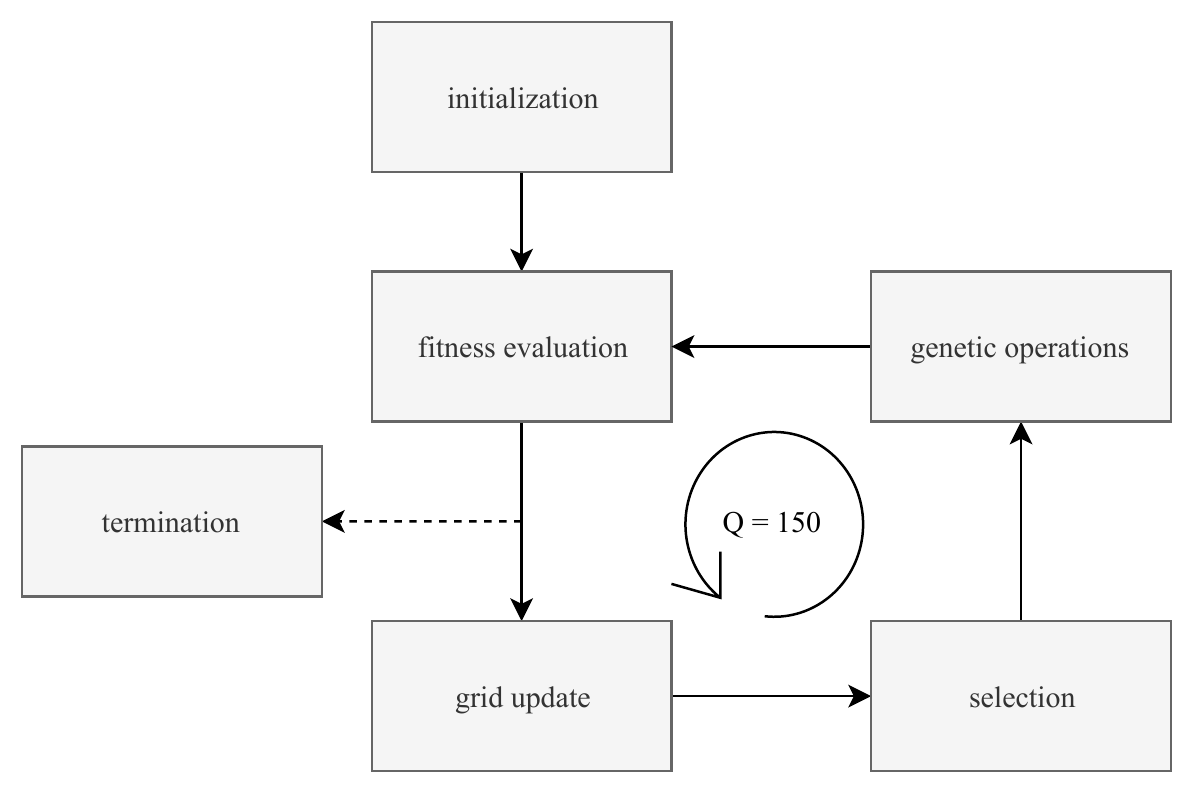}
        \caption{High-level view of MAP-Elites iterations.\label{fig:map}}
	\end{centering}
\end{figure}

The algorithm is then quite simple. After the generation of initial random expressions, individuals are evaluated. The individuals produced are then sorted in terms of grid bins. Inside a given bin, only the best individual is kept. Once the grid has been populated, an iteration consists in producing new equations by applying genetic operators between the elements of the grid -- and only them -- chosen at random (uniform selection, as in \cite{DBLP:journals/corr/MouretC15}). These new states are then evaluated, and the grid is updated: some of these new equations may replace some previous best individuals in several bins of the grid, see Fig. \ref{fig:map}. Termination is similar to the pure GP case.

The algorithm, and its relation to other standard evolutionary methods such as, \eg{} Pareto-Front optimization or NS-LC is discussed at length in \cite{DBLP:journals/corr/MouretC15}. We want to emphasize that using such a feature map shall address in many ways the aforementioned issues of GP for SR. First of all, the preservation of diversity is kind of built-in in the method, while addressing at the same time the quest for multi-objective regression. Secondly, the bloat can be addressed by choosing as a feature the length of an equation (or its complexity). This will indeed force the algorithm to remember small individuals and automatically counter the bloat. Small individuals may not be excellent ones, but still are the best seen so far of that given size or complexity. As such, it also acts as a kind of problem decomposition, where small individuals can be seen as relevant blocks for later building a larger and better equation. Regarding now the parsimony and Occam's razor, it is clear that when two individuals have similar fitness, the one using less free scalar should be considered as better than the other one. Therefore, it is natural to use as a feature the free scalars count of the equation. On top of increasing the diversity, it provides a way to rank the best solutions by the number of free parameters used, which is good practice when dealing with analytic models of physical phenomena (compare to approach of \cite{de2018greedy}).

We believe that using both length (or complexity), and the number of free scalars in an equation, are two inescapable choices in MAP-Elites-based Symbolic Regression. Additional features might be included, either to increase the grid size, or to bring some more relevant diversity. To determine which additional features to use is quite a non-trivial question. We have chosen a 3-D grid (see Section \ref{sec:model}) based on the length, the number of scalars, and the number of non-algebraic functions such as $\sin$ or $\exp$. This last choice is arguable, but does increase the grid size and thus boosts the exploration.

\subsection{CMA-ES}
\label{sec:background:cmaes}
CMA-ES method is described at length in a series of papers, see, \eg{}, \cite{hansen2003reducing} and available as a library in many programming languages. Although details are quite complex, the main idea is that the algorithm will browse the landscape in an evolutionary way. Say one wants to minimize a function $f(A_1, \ldots, A_n)$. First, a population of candidates for $\vec{A}$ are sampled along a normalized multivariate Gaussian; then the best individuals (in a mean-squared error sense with respect to the actual target) are sub-sampled, and this in turn defines a new multivariate normal with a shifted mean and covariant matrix, according to which new candidates are drawn, \etc{}. The method requires many such iterations (from \SI{1000}{} to \SI{10000}{}), and each iteration relies on some number of function evaluation resulting in a quite slow, but powerful method. It can be trapped in local optima, of course, but is also able to capture quite often the global optimum. See also Section \ref{sec:results:grid} for an explicit example. Meta-parameters include, amongst other things, the population size, the maximum number of iterations, and a time limit that we have modified with respect to default values -- see next section for details. 

As an example, consider the target "Korns-7" (as named in \cite{DBLP:journals/corr/abs-1805-10365}, \ie{} $f_{\textrm{target}} = 213.80940889 \,(1 - e^{-0.54723748542 \, x})$ on the range $x \in [-50, 50]$. It is formally quite a trivial target, but is also understandably difficult to find exactly without an appropriate method for finding these two scalars. Here CMA-ES does trivialize finding such an equation. In fact, such a simple equation is likely to be already present in the initial random population. In that case, applying then the CMA-ES method to find the best-fit for the $A$'s will terminate the algorithm in only one step. As a matter of fact, the target Korns-7 was always found very easily. Also, because of its triviality for our combined method MAP-Elites + CMA-ES, we decided to remove it from our target list.

\section{The model}
\label{sec:model}
\subsection{General specifications}
\label{sec:model:general}
The model shall run on \emph{all the targets} of Tables 1--5 with the \emph{same} following primitive set:
\begin{equation}
\label{voc}
    \left \{\varnothing, x, y, z, \sin, \cos, \exp, \ln, \times, +, -, /, \wedge \right \}
\end{equation}
where $\wedge$ stands for exponentiation. We limited the set to three variables ($x ,y, z$) at most for reasons to be discussed later. The dictionary is then completed by pure numbers (also referred to as scalars in the following). Then two options exist. First the model can be run with integer scalars (namely '1' and '2' only in the following): this model is used to compare plain GP to MAP-Elites SR, see Section \ref{sec:model:comparison}. Second, the full model can be run with free floating-point scalars $A$ that are fitted by the CMA-ES method at the end, as detailed in Section \ref{sec:model:full}.

Basic simplifications on the fly are also included, such as $\exp(\ln(x)) =x$, \etc{}. As we do not rely on existing simplification packages, the algorithm is not equipped with a full expand-refactor simplification engine. Instead, it is limited to a set of basic hard-coded rules, which is sufficient for the application described in this paper. Note that when using formal scalars $A$, an expression like $A \times A \times \exp(x)$ can be simplified to $A \times \exp(x)$ prior to CMA-ES evaluation. Some simplifications require to add some special symbols to our dictionary, namely the neutral element -- if not already present--, the zero, and infinity.

We do not use protected divide of any kind. When infinity occurs in simplifications such as $1/(x -x) \to 1/0 \to \infty$, the equation is discarded prior to evaluation. Because our simplification engine is not comprehensive however, the program occasionally creates zero division errors, in which case the maximum penalty is attributed to the equation. The same applies for other kind of exceptions such as overflow errors.

We restrict ourselves to very basic genetic operations, namely point-wise mutations between symbols of equal arities, and basic crossovers. By this we mean nodes for crossovers are chosen randomly amongst internal nodes and leaves. We do not try to improve this by weighting probabilities for choosing internal versus terminal nodes, for instance, or other refinements such as the ones described in \cite{poli2008field}. Because we set a hard limit on the length of an equation, crossovers are tried but discarded as soon as the resulting equation does not fit inside this limit. This is a bit different from the literature where usually a hard limit on max depth of the tree is set up (using a maximum length $L$ instead is more relevant when using a MAP-Elites grid).

Also, and this can be seen as the only expert knowledge we do implement, we limit ourselves to a maximum number $K$ of nested functions such as $\sin(\sin(\exp(...)))$ for interpretability reasons. In particular, runs were made with $K=1$ ($K=2$ also works fine). Again, crossovers leading to out-of-bound equations are discarded. Alternatively, the number of nested functions could have been used as yet another feature for the MAP-Elites grid.

Regarding genetic operations, states are chosen randomly among the population (both for plain GP and MAP-Elites), and in $40\%$ of the cases, one symbol only is mutated, in $40\%$ of the cases, two random states go through a crossover and returns two offspring, and in the last $20 \%$, two states are chosen at random and go through both a mutation and a crossover. Also, when a non-algebraic function is chosen for mutation, there is a $30 \%$ probability to simply drop the symbol as in $x\times\sin(x) \rightarrow x^2$. This choice was made in order to limit the number of non-algebraic functions and helps fighting the growing number of nested functions that crossover usually produce (unless hard limit on such terms is set).

As said in Section \ref{sec:background}, the MAP-Elites grid that we use is three dimensional and uses as features the length of the equation (in post-fix notation), the number of free scalar parameters, and the number of non-algebraic functions such a $\sin, \exp, \ldots$ involved in an expression. We thus use a grid with one bin for each length of the equation from $1$ to $L$, one bin for each number of any non-algebraic function from $0$ to $8$ (if an equation has more than $8$ functions, then it enters the last bin), and one bin for each free scalar from $0$ to $L/2$ (which is the maximum possible). In order to give orders of magnitude, the grid is usually small with around $150$ elements for equations with a maximum size of $L=15$, and may be as large as around a thousand individuals for larger equations with $L \geq 35$. 

Finally, because CMA-ES is not a perfect method, once it has returned a recommendation for the best $A_i$s, we apply thereafter a least square method to descend to the closest minimum, if not exactly found previously. This usually increases the method's precision. As a result, for a (trivial) target like, say, $f=x^3$, our method might well return the exact result $f = x \times x \times x$ or, also, $f=x^A$. In the latter case, the CMA-ES plus the least-square fit combined will in general return $A = 2.99999(\ldots)$. In this case, we consider that the target has been exactly found, see next section for termination criterion. Note that because non-integral exponents are permitted, we need to restrict the sampled values for the variables to positive ranges, see Section \ref{sec:results:targets}.

We have implemented the model in \textsc{Python}, using \textsc{CMA-ES for Python} \cite{cmapackage}, and \textsc{scipy} module for least squares. Everything else has been coded from scratch.

\subsection{Cost function and termination criterion}
\label{sec:model:termination}

Following common practices of SR literature, the program write equations $\tilde{f}= \tilde{f}(\textrm{primitive set})$, where the right-hand side cannot\footnote{This prevents finding polynomial equations in $f$ that would first require a numerical solver which is a more complex task and left for future work.} have terms depending on $\tilde{f}$. Differential equations of the form $(d/dx)^n \tilde{f} = \ldots$ could also be produced using the same approach, but we restrict ourselves to non-differential equations in this paper.

Once the best free scalars $A$ in $\tilde{f}$ are found, we simply compare the training set's right-hand side with the target. Similarly of what can be found in multi-objective regression papers, we found more effective to also take into account a "derivative cost". We thus use the cost $C$:
\begin{equation}
\label{cost}
C = \vert f_{\textrm{target}}-\tilde{f}\vert + \sum_i \vert \partial_i (f_{\textrm{target}} - \tilde f)\vert 
\end{equation}
\ie{}, using the L1 norm on both the distance to the target, and the distance of the first derivatives to the target. A sum is used for the derivative cost when the function has more than one variable. The cost is then properly normalized in order to return a reward (or fitness) scaled between $-1$ and $1$. By running some preliminary tests, we found that including the derivative cost speeds up the convergence. 

Because CMA-ES is so accurate however, termination criterion can be subtle. Indeed, the program quite often produces spectacular fits (accurate to $10^{-5}$ in relative values) to the target on its \emph{training} set, even if the formal equation is not the expected one. See for instance Fig. \ref{fig:fit}. Therefore, in our target list in Section \ref{sec:results} taken from the literature, we have been careful to often increase the range of the \emph{validation} set to prevent the algorithm from stopping early on false positives. (We recall that termination criterion is based on the validation set only). Going back to the trivial example where the target is $f_{\textrm{target}}=x^3$, our method might return $f=x^A $ with $A = 2.99999(\ldots)$. We consider that the target is exactly found in this case, in the following sense: one computes the NRMSE\footnote{As defined in Eq. (6) in \cite{miranda2017noisy}.} (Normalized Root Mean Square Error) which in that case would be typically close to $10^{-16}$ due to numerical precision. We defined our termination criterion as $NRMSE \leq 10^{-6}$.

\subsection{Comparison of plain GP versus MAP-Elites}
\label{sec:model:comparison}
As said, we made some preliminary runs to check whether MAP-Elites improves GP approaches. In order to do so, we used a vocabulary with two integer scalars '1' and '2' (\ie{} no CMA-ES), on targets of Table 1. Genetic operations have the same parameters in both runs; GP runs with a population size of 1000 individuals. Selection is made with a tournament\footnote{Two individuals are drawn randomly from the population. Only the best one is kept for mutation, otherwise they go through a crossover.} of size 2, and genetic operations are done until 2000 new individuals are produced. These ones are then evaluated, and the 1000 bests of these 3000 individuals become the new population. Regarding the MAP-Elites run, 4000 individuals are first randomly created, evaluated, and binned in the grid. Then, when the grid has $N$ elements, $2N$ new individuals are created by genetic operations, evaluated, binned, and the grid is updated. 

The program either stops when the target is hit, or after $10^5$ evaluations of the fitness function. We made 100 such runs for the five targets of Table 1. It shows that MAP-Elites is a slight improvement over GP although it requires slightly more fitness evaluations.

\subsection{Full algorithm MAP-Elites + CMA-ES}
\label{sec:model:full}
We could have simply used a MAP-Elites + CMA-ES algorithm on an initial collection of random equations. Literature often goes for the half/half method for initializing the population. In fact, as we have realized that CMA-ES is a slow method while using integer scalars is very fast, our full algorithm is rather three-steps:

\begin{itemize}
    \item First, we create a collection of 4000 random equations. Then we evolve for $P = 150$ iterations the grid of equations by using a grid with no free scalar $A$, but only scalars "1" and "2". The maximum size is set to $L +10$. In the case where the target includes no floating-point number, it may already be found at this step, and quite often is. This is step 1 of Fig. \ref{fig:fullalgo}. 
    
    \item If not, promising equations from the grid such as $\tilde{f} = \sin(2) x/(1+x)$ are transformed into their free scalar counterpart, namely: $\tilde{f} = \sin(A_1) x/(A_2+x)$, \ie{} $\tilde{f} = A_1 \ x/(A_2+x)$ after simplification. This is step 2 of Fig. \ref{fig:fullalgo}. They are then evaluated by CMA-ES, and stored inside a new grid. 
    
    \item This grid serves as an initialization for $Q=150$ iterations using MAP-Elites with free scalars $A$ and CMA-ES, see step 3 of Fig. \ref{fig:fullalgo}.
    
\end{itemize}

In other words, we use the MAP-Elites method with ephemeral constants (drawn randomly from the integer set $(1,2)$) to generate an initial population for CMA-ES which is \textit{a priori} much more relevant than random equations. This means that we do have two distinct vocabularies and two distinct sets of simplification rules in our code.

The full algorithm can be summarized by the following diagram:
 \begin{figure}[ht]
	\begin{centering}
		\includegraphics[width=\linewidth]{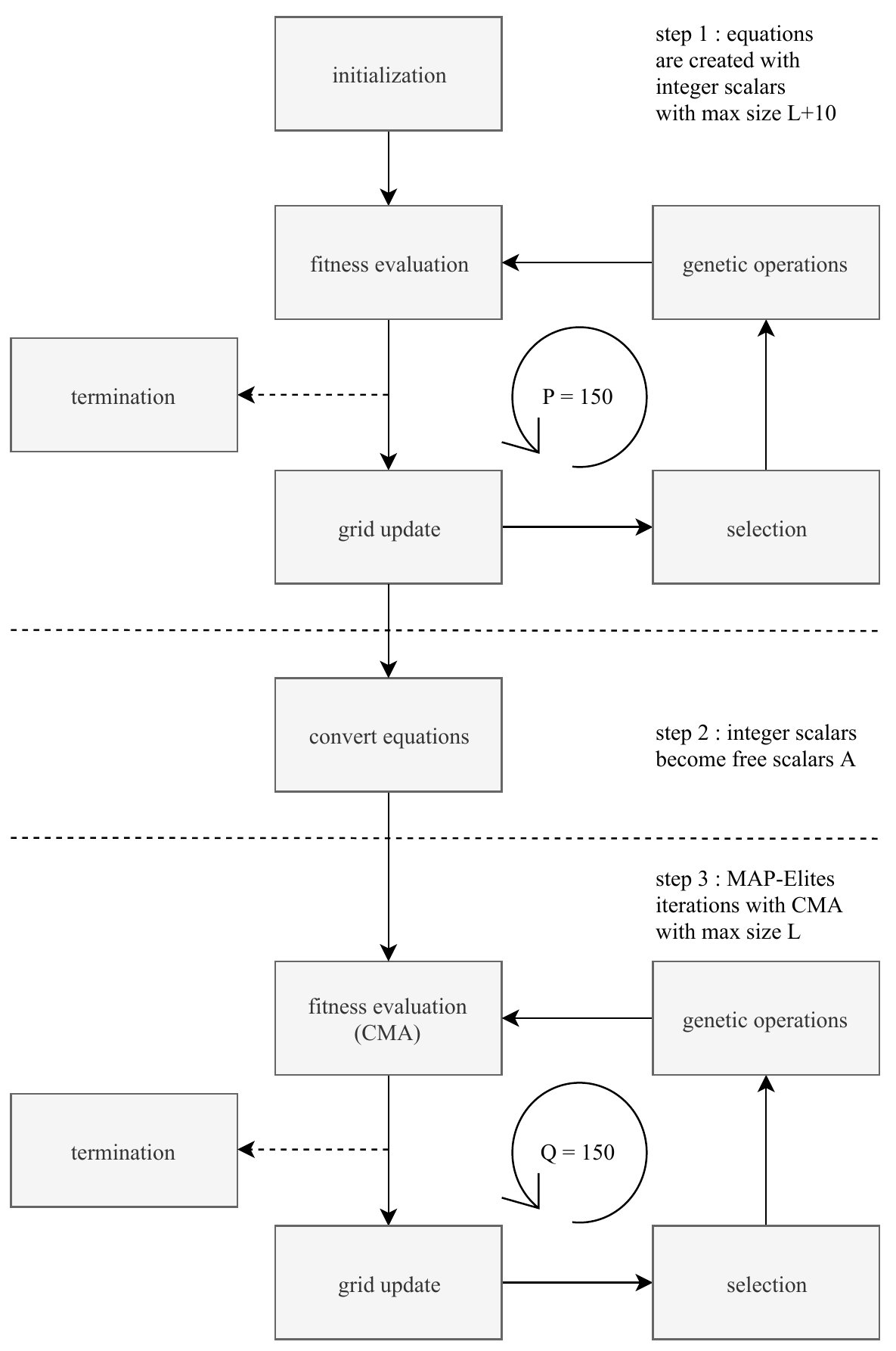}
        \caption{High-level view of the full algorithm.\label{fig:fullalgo}}
	\end{centering}
\end{figure}

 As a remark, even if it is not required in a strict sense, simplification often reduces the number of free parameters per equation. For instance $\tilde{f} = A\times A\times A\times(x+A)$ is equivalent to $\tilde{f} = A_1\times x + A_2$. Being able to make this simplification is a huge advantage because the fewer free scalars, the faster CMA-ES method executes. Next subsection gives order of magnitude about execution time.

\subsection{Execution time}
\label{sec:model:execution}
The first step of the algorithm described above runs way faster than the second one, even in mono-CPU implementation. The execution time is in fact no different than the one of standard SR packages like DEAP \cite{DeRainville:2012:DPF:2330784.2330799}.

The CMA-ES method is very much time-consuming. On mid-sized equations (say of length $30$) with around 5 free scalars, the CMA-ES method takes around 30 seconds on a single CPU. In practice, we ran CMA-ES on a 40 cores machine. But even in this case, for difficult targets with maximum size $L=40$ or $45$, it took almost two days to produce around \SI{200000}{} CMA-ES evaluations. The runs we report in Table 5 for difficult targets typically took between one and three days per target per run. Clearly, our method can not be generalized to very long expressions.

The details of the execution time required for each run are reported in the captions of Tables 1 to 5.

\section{Experimental Results}
\label{sec:results}
\subsection{Targets description}
\label{sec:results:targets}
As already said, we use the same vocabulary for all our targets. The only change from one to another is the maximum length an individual can have. The table of targets we used can be found on the last page of Ref.~\cite{DBLP:journals/corr/abs-1805-10365}, which is in itself a compilation of many targets from the literature (see also \cite{white2013better}). From this original list of 63 targets, we discarded the ones with more than three variables as well as the ones that are too simple for our implementation\footnote{Namely Keijzer-7, Keijzer-8, Keijzer-13, Korns-4, Korns-5, Korns-6, Korns-7, Nguyen-1, Nguyen-8, Vladisleva-6}. 

We ended up with the 31 targets that are listed in result Tables 1 to 5. Table 2 lists "easy targets" for which a sentence length of 15 was enough (although the success rate might improve with slightly longer maximal size), while Table 3 reports "reasonably difficult ones" with length around $L=30$, and Table 4 and 5 list difficult targets for which our success rate dropped significantly. 

As in Ref.~\cite{DBLP:journals/corr/abs-1805-10365}, the notation $x : E[0,1, 0.05]$ means that the variable $x$ is sampled in the interval $[0,1]$ with constant steps of size $0.05$, while $x : U[0,1, 100]$ means for instance sampling 100 points for $x$ in the interval $[0,1]$ according to uniform distribution.

Our success rate is in most cases greater than $80 \%$. One might object that fixing the maximal length $L$ from the start is expert-knowledge. But in practice $L$ is just a meta-parameter that can be adjusted by the user, starting from small values that lead to shorter execution times, and progressively increasing it when the success rate is too small. As a matter of fact, we did proceed in this way for some of the targets of Table 3 that we first thought would be easy in small length, but turned out to be harder than expected. In theory, nothing would prevent a more generic version of the algorithm presented in this paper to auto-adjust this parameter.

Since the CMA-ES method process floating-point numbers, one must avoid expressions that are not defined on $\mathbb{R}$, such as $x^{2.999}$ for $x\in\mathbb{R}^{-}$. 
Therefore, compared to the table \cite{DBLP:journals/corr/abs-1805-10365}, the training and validation set ranges were systematically cut to subsets of $\mathbb{R}^+$. The same number of points were given for the training sets, but on a \emph{smaller} range. Yet, it did not prevent us to achieve high success rates.

Also, we give ourselves a much smaller training set for some of the multidimensional targets. Consider for example Keijzer-5 (see Table 2). Ref.~\cite{DBLP:journals/corr/abs-1805-10365} reports a training set $U [-1, 1, 1000]$ for $x, y$ and $U [-1, 1, 10000]$ for $z$. This would mean providing way too many points to the CMA-ES method, which already requires many iterations. This would translate into a very large execution time. For this reason, we limited the training set for targets of this sort. In particular, for this target, we give ourselves a training set of $5\times 5\times 10$ points only. Again, this does not prevent the method to achieve $100 \%$ success rate here on the validation set.

However, this is also a clear drawback of the method for high dimensional targets. Remind that CMA-ES optimizes the $A_i$'s in $\tilde{f}(\vec A, \vec x)$ with respect to the quadratic cost $\sum_{\vec x}(f_{\textrm{target}}(\vec{x}) - \tilde{f}(\vec A, \vec x))^2$. Therefore it can not really do so without taking too long when $\vec x$ is of the order of a thousand points. Thus, in practice, we could only experiment with this method up to 4-dimensional targets with 5 points along each dimensions ($5^4 = 625$). For this reason, we explore at most three-dimensional targets in this paper. This limitation may however be overcome by using another optimizer than CMA-ES.

\subsection{Result tables description}
\label{sec:results:tables}
Table 1 is self explanatory. Regarding results reported on Tables 2, 3, and 5 of the combined method MAP-Elites + CMA-ES of Section \ref{sec:model:full}, the first column corresponds to the target name listed in \cite{DBLP:journals/corr/abs-1805-10365}. The second column gives the formula, while the third column details the training set. As said, we only reduced the range to positive values with respect to \cite{DBLP:journals/corr/abs-1805-10365}, and sometimes we reduced the number of points provided to the evolutionary algorithm, but never increased them. The fourth column corresponds to the validation set, usually greater than the one given in this reference, for reasons already explained in Section \ref{sec:model:termination}.

Then, the fifth column gives the hit rate for the first step of the algorithm with integer scalars $1$ and $2$. Given that only a few of these targets involve floating-point numbers, this first step is already able to reach the target, especially the easy ones of Table 2. On the contrary, it can never hit Keijzer 1,2,3 which are $0.3 x \sin(2 \pi x)$ on various training sets. Column 6 gives the number of evaluations that were done before actually hitting the target, and when it was hit, averaged over the number of independent runs.

Column 7 and 8 are similar to 5 and 6, but this time for the second step of the algorithm involving CMA-ES, and after conversion of integer scalars to free scalars $A_i$. The last column is the sum of both hit rates, \ie{} our main experimental result.

The following Tables have only 29 targets, and not 31: this is because we shall not report on Vladislavleva-7 $$f = (x-3)(y-3) + 2 \sin((x-4) (y-4))$$ and Vladislavleva-5 $$f = \frac{30 (x-1) (z-1)}{(x-10) y^2},$$ for which we have no success at all (although quite good NRMSE).

\begin{table*}[p]

\caption{Comparison between plain GP and Map-Elites with only free scalars being ``$1$'' and ``$2$''. See Section \ref{sec:model:comparison} for more details. Based on 100 independent runs. $N$-eval is the average number of individuals evaluated when the solution \emph{is} exactly found. MAP-Elites shows a slight improvement over plain GP, although it requires a bit more evaluations before convergence. The training and validation set intervals are the same as the ones specified in Table 2.\label{Table1}}
\begin{tabularx}{\textwidth}{@{}l*{10}{c}c@{}}

\toprule
Target name &  Target formula & Hit rate - GP & N-eval & Hit rate - Map-Elite & N-eval  \\ 

Nguyen-2    & $x + x^2 + x^3 + x^4$   &67 \%     & 22292 &  93 \%   & 28969 \\

Koza-3    & $x^6 -2 x^4 + x^2$     & 24 \%    & 34454  & 50 \% & 42143  \\

Meier-3*    & $x^2 y^2/(x+y)$     &94 \%     & 29251 & 100 \% & 26320 
   \\ 
   
Meier-4*    & $x^5 y^{-3}$     &57 \%  & 41364    &52 \%  & 59896
     \\ 

Nguyen-9*    & $\sin(x)+ \sin(y^2)$  &  87 \%  & 20121 &85 \% & 56461  
   \\ 

Burks    & $4 x^4 + 3 x^3 + 2 x^2 + x$   & 2 \% & 46881   &34 \% & 74735

\\
\bottomrule
\end{tabularx}

\end{table*}

\begin{table*}[p]
 \caption{Results for small targets with $L=25$ for the first step (no CMA-ES) and $L=15$ with CMA-ES. Starred targets correspond to targets where the intervals for $x$, $y$ (and $z$ if any) are the same. Based on 20 independent runs. When the run fails after $P = 150$ iterations for the first step and $Q = 150$ iterations for second step, computation time is around 20 minutes on a 40-cores computer (for one target).\label{Table2}}
\begin{tabularx}{\textwidth}{@{}l*{10}{c}c@{}}
\toprule
Target name & Target formula  & Training set & Test Set & Hits - no CMA-ES & N-eval & Hits (CMA-ES) & N-eval & Hits (total)  \\ 

Nguyen-2    & $x + x^2 + x^3 + x^4$   & U [0, 1, 20]   & U [0, 2, 200]   &95 \%     & 24624 &  5 \%  & 4221  & \textbf{100 \%} \\

Koza-3    & $x^6 -2 x^4 + x^2$   & U [0, 1, 20]   & U [0, 2, 200]   &40 \%    & 48722 & 45 \% & 16899 & \textbf{85 \%} \\

Meier-3*    & $x^2 y^2/(x+y)$   & U [0, 1, 20]   & U [0, 2, 50]    &100 \%     & 27948 & -  & - & \textbf{100 \%}
   \\ 
   
Meier-4*    & $x^5 y^{-3}$   & U [0, 1, 20]   & U [0, 2, 50]   &80 \%  & 41217    & 20 \% & 3957 & \textbf{100 \%}
     \\ 

Nguyen-9*    & $\sin(x)+ \sin(y^2)$   & U [0, 1, 20]   & U [0, 2, 100]   &90 \% & 56254 & 10 \% & 2184   & \textbf{100 \%}
   \\ 

Keijzer-1    & $0.3 x \sin(2\pi x)$   & E [0, 1, 0.05]   & E [0, 10, 0.05]   &0 \%  & - &95 \% & 5704   & \textbf{95 \%}
  \\ 

Keijzer-2     & $0.3 x\sin(2 \pi x)$   & E [0, 2, 0.05]   & E [0, 4, 0.05]   &0 \% &  - & 100 \% & 5611     & \textbf{100 \%}
  \\ 

Keijzer-3    & $0.3 x \sin(2 \pi x)$   & E [0, 3, 0.05]   & E [0, 4, 0.05]   &0 \% &  - & 100 \% & 3717     & \textbf{100 \%}
 \\ 
Nguyen-5     & $\sin(x^2) \cos(x) -1$   & U [0, 1, 20]   & U [0, 1.2, 200]   &20 \% &  46194 & 60 \% & 19551   & \textbf{80 \%}
  \\ 
Nguyen-6    &  $\sin(x) + \sin(x + x^2)$   & U [0, 1, 20]   & U [0, 1.2, 200]   &60 \%     & 48362 & 35 \% & 13898 & \textbf{95 \%}
\\

Sine    &  $\sin(x) + \sin(x + x^2)$   & E [0, 6.2, 0.1]   & U [0, 10, 100]   &90 \%  & 34619   & 10 \% & 10417 & \textbf{100 \%}
\\

Koza-2    & $x^5 -2 x^3 + x$   & U [0, 1, 20]   & U [0, 2, 200]     &45 \%  & 57392   & 50 \% & 17520 & \textbf{95 \%}

\\
\bottomrule
\end{tabularx}
\end{table*}

\begin{table*}[p]
\caption{Results for mid-sized targets. Based on averaging 20 independent runs. Burks, Keijzer-14 and Nguyen-3 are run with maximum $L$ of 20 for the CMA-ES method, Nguyen-7 with $L=25$, and the remaining ones with $L=30$. One run per target takes at most one hour for $L=20$ (\ie{} when it fails), and up to 5 hours for $L=30$.
\label{Table3}}
\begin{tabularx}{\textwidth}{@{}l*{10}{c}c@{}}
\toprule
Target name & Target formula  & Training set & Test Set & Hits  & N-eval & Hits  & N-eval & Hits   \\ 
 &  &  & &(no CMA-ES) &  &(CMA-ES) & & (total)
\\

Burks    & $4 x^4 + 3 x^3 + 2 x^2 + x$   & U [0, 1, 20]   & U [0, 3, 200]    &35 \% & 79033 & 60 \% & 16350   & \textbf{95 \%}
\\
Keijzer-14*    & $ \frac{8}{2 + x^2 + y^2}$   & U [0, 3, 20]   & E [0, 4, 0.1]   &30 \%  & 139554     & 65 \% & 6644 & \textbf{95 \%}
\\
Nguyen-3    & $x + x^2 + x^3 + x^4 + x^5$   & U [0, 1, 20]   & U [0, 2, 200]   &60 \%  &65082   & 30 \% & 17523 & \textbf{90 \%}
\\
Nguyen-7 & $\ln(1+x) + \ln(1+ x^2)$ & U [0, 2, 20] & U [0, 3, 200]  &0 \%  & -    & 20 \% & 42459 & \textbf{20 \%}
\\

R1    &  $(x+1)^3/(x^2 - x +1)$   & E [0, 2, 0.1]   & U [0, 3, 100]   &5 \%  &143850    & 90 \% & 50741 & \textbf{95 \%} 
\\
R2    &  $(x^5 - 3 x^3 +1)/(x^2 +1)$   & E [0, 2, 0.1]   & U [0, 4, 400]   &0  \%  & -    & 85 \% & 73009 & \textbf{85 \%}
\\
Keijzer-5   & $30 x z/((x-10) y^2)$   & $\frac{x,y \, : \, U [0, 2, 5]}{z \, :  \, U [1, 5, 10]}$    & $\frac{x, y \, : \, U [0, 3, 20]}{z \, : \, U [0, 5, 30]} $  &5 \%  & 348979    & 95 \% & 14983 & \textbf{100 \%}
\\
Keijzer-12*  & $x^4 - x^3 + 0.5 y^2 - y$   & U [0, 3, 20]   & E [0, 4, 0.1]   &30 \%  & 259639     & 70 \%
& 47086 & \textbf{100 \%} \\

Keijzer-15*    & $\frac{x^3}{5} + \frac{y^3}{2} - y - x $  & U [0, 3, 20]   & E [0, 4, 0.1]   &0 \% & - & 100 \% & 35894     &\textbf{100 \%}
\\
Keijzer-11    & $x y + \sin((x-1)(y-1))$   & U [0, 3, 20]   & E [0, 4, 0.1]   &0 \%  & -    & 15 \%
 & 71605 & \textbf{15 \%}\\
 
Nguyen-4    & $x + x^2 + x^3+ x^4 + x^5 + x^6$   & U [0, 1, 40]   & U [0, 1.5, 200]   &40 \% & 181816  & 55 \% & 41728  & \textbf{95 \%}
\\
Pagie-1    &  $1/(1 + x^{-4}) + 1/(1+ y^{-4})$   & E [0, 5, 0.2]   & U [0, 6, 20]   &15 \%  & 233542 & 85 \% & 48647    & \textbf{100\%}
\\

\bottomrule
\end{tabularx}
 
\end{table*}

\begin{table*}[!htpb]
\caption{Description of difficult targets. $\quad$ $\quad$  $\quad$ $\quad$ $\quad$ $\quad$ $\quad$ $\quad$ $\qquad$  $\qquad$ $\qquad$ $\qquad$ $\qquad$  $\qquad$ $\qquad$ $\qquad$ $\qquad$ $\qquad$ $\qquad$ $\qquad$ $\qquad$\label{Table4}}
\begin{tabularx}{\textwidth}{@{}l*{10}{c}c@{}}
\toprule
Target name & Target formula  & Training set & Test Set & Maximal Length used (CMA-ES)\\

R3    &  $\frac{x^6 + x^5}{x^4 + x^3 +x^2 + x +1}$   &  E [0, 1, 0.05]   & U [0, 2, 100]  &  35
\\
Vladislavleva-1    & $e^{-(x-1)^2}/((1.2 + (y-2.5)^2)$   & (x,y) : U [0.3, 4, 20]   & (x,y) : E [0, 8, 0.1]   & 35
\\
Keijzer-4    & $x^3 e^{-x} \cos(x) \sin(x) (\cos(x) \sin(x)^2-1)$   & E [0, 10, 0.1]   & U [0, 14, 1000]  & 40
\\
Nonic    & $\sum_{i=1}^{i=9} x^i $   & E [0, 1, 0.05]   & U [0, 2, 100]  & 40
\\
Vladislavleva-3    & $x^3 e^{-x}\cos(x)\sin(x)(\cos(x)\sin(x)^2-1)(y-5)$   & x : E [0.05, 10, 0.1]   & x : U [0, 10, 50] & 45
\\
  &    &  \, y :  E [0.05, 10.05, 2]   &  y : U [0, 10, 10]
\\
\bottomrule
\end{tabularx}

\end{table*}

\begin{table*}[!htpb]
\caption{Results for difficult targets, based on 10 independent runs. The first step of the algorithm (without CMA-ES) can see more than \SI{500000}{} different equations. This is however not enough to hit the target. Computational time is around two days per target per run. \label{Table5}}
\begin{tabularx}{\textwidth}{@{}l*{10}{c}c@{}}
\toprule
Target name & Hits - no CMA-ES & N-eval & Hits (CMA-ES) & N-eval & Hits (total)  \\ 
R3    &20 \% & 325312 & 70 \% & 64319 &\textbf{90\%}
\\
Vladislavleva-1       &0 \%  & -    & 30 \%   & 101360  & \textbf{30 \%}
\\
Keijzer-4     &0 \%     & -   & 40 \%  & 105969 & \textbf{40 \%}
\\
Nonic         &0 \%  & -   & 20 \%  & 170928 & \textbf{20 \%}
\\
Vladislavleva-3       &0 \% & -  & 20 \%  & 246756   & \textbf{20 \%}
\\

\bottomrule
\end{tabularx}
 
\end{table*}

\subsection{Illustration of MAP-Elites Grid}
\label{sec:results:grid}
As an illustration, next Fig. \ref{fig:grid} shows the population on the MAP-Elites grid for a successful run for target 'Burks', as described in Table 1 and 2. The $x$-axis corresponds to the number of free scalars from 0 to 16 and $y$-axis is the length (from 0 on the top to 30). This is not the full 3-D grid, but only a slice corresponding to a number of functions less or equal to 1. The grid is not (and actually can't be) populated on the top right corner. It shows that small individuals perform quite poorly, as well as individuals with too many or too few free scalars.

\begin{figure}[!htpb]
	\begin{centering}
	    \vspace{-\baselineskip}
		\includegraphics[width=\linewidth]{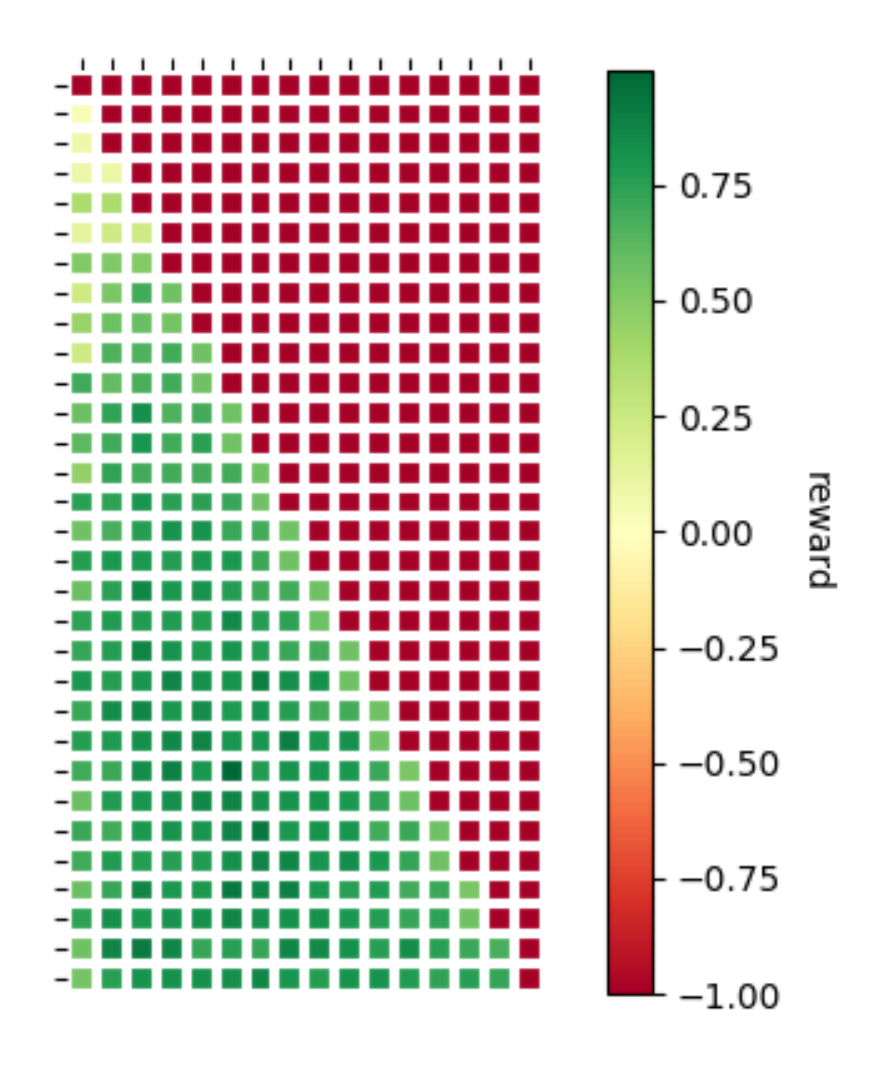}
		\vspace{-\baselineskip}
		\caption{A slice of the MAP-Elites grid for target Burks $4 x^4 + 3 x^3 +2 x^2+x$.\label{fig:grid}} 
	\end{centering}
\end{figure}

\subsection{Discussion}
\label{sec:results:discussion}
\begin{figure}[!htpb]
	\begin{centering}
	    \vspace{-\baselineskip}
		\includegraphics[width=\linewidth]{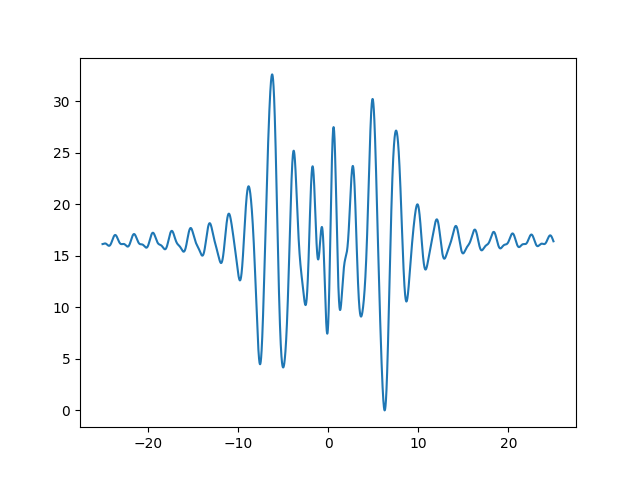}
		\vspace{-\baselineskip}
        \caption{Landscape projection on only one dimension : this is $g(a) = \sum_{x_i}\left(0.3 x_i \sin(2 \pi x_i) - 0.3 x_i \sin(a x_i) \right)^2$ where $x_i$ are given by the training set in Table 2, Keijzer-3.\label{fig:mse}} 
	\end{centering}
\end{figure}
\begin{figure}[!htpb]
	\begin{centering}
	    \vspace{-\baselineskip}
		\includegraphics[width=\linewidth]{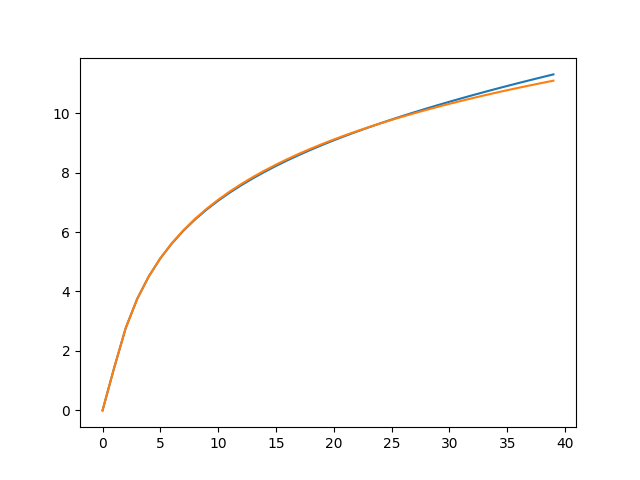}
		\vspace{-\baselineskip}
		\caption{In blue, the function found by the program, Eq. (3) versus the target $f = \ln((1+x)(1+x^2))$ on the range $x \in [0,40]$. Note that it has been trained only on the range [0,2].\label{fig:fit}} 
	\end{centering}
\end{figure}

It appears that target involving $sin$ functions are the most difficult ones. This is can be illustrated by Keijzer 1, 2, 3, $f = 0.3 x \sin(2 \pi x)$ that require many fitness evaluations before convergence although the target looks quite trivial (and see also the low hit rate for target Keijzer-11). As a matter of fact, the CMA-ES method has a very small success rate on fitting the exact equation $\tilde{f} = A_1 x \sin(A_2 x)$. The success rate with default CMA parameters is only around $2 \%$ based on 1000 runs for Keijzer-1. This means that CMA-ES will often miss the right target. 

This is because the landscape is quite deceptive in this case. Next figure shows projections of that landscape as a function of $A_2$ (on the horizontal axis) alone of the mean squared error (used in the CMA-ES method) between the actual target and a test function $f = 0.3 x \sin(A_2 x)$ for the test range of Keijzer-3. We see that the global minimum in $A_2 = 2 \pi$ is kind of lost in many local optima, see Fig. \ref{fig:mse}. 

In fact, when the target is found, it is usually with a more complex formula, \eg{} $\tilde{f} = A_1 x \sin(A_2 x + A_3)$ or even $\tilde{f} = A_1 x \sin(A_2 x + A_3) +A_4$, for which CMA-ES success rate increases significantly up to around $17 \%$. Interestingly enough then, the algorithm tries many variations around the correct formal expression until CMA-ES hits the right scalar parameters. Moreover CMA-ES seems to be able to do better on landscapes that have extra spurious dimensions. Most presumably, the landscape gets easier to browse in this extra dimensional embedding. This also means that simplification may be counter-productive. However, we did runs with and without simplifications and the results are overall similar. (Simplification is not used from Table 1 to 4, and used in Table 5).

In order to increase CMA-ES success rate, each CMA-ES instance is called with an initial Gaussian with a random mean varying between $-1$ and $1$, and a random initial variance chosen varying between 1 and 5. 

The poor success rate on target Nguyen-7 (Table 3) is of different nature. It seems, looking at the detailed result, that the target on that range can be very easily approximated by rational fractions, so that the program does not get incentives to look for $\log$ solutions. Moreover, the provided range is very small; the success rate increases a lot and reaches easily $100 \%$ if we give a training set of \eg, $x \in [0,40]$. Given that we give ourselves only a small sample of the function, the difficult task here is more about finding good generalizations. In this regards, the algorithm actually performs very well. As an example, one of the program's result is the following (with a NRMSE of $5.6 \, 10^{-5}$):
\begin{eqnarray}
\tilde{f} &=& 0.000219974 \\ \nonumber &+&\frac{0.562568 x \left(12.9747 + x^{0.593097} - \frac{8.99871}{x^{2.30042} + x^{1.17464}+1.343}\right)}{x + 3.57942} 
\end{eqnarray}
which is highly accurate not only on the validation set $x \in [0,3]$ but actually on a much larger range $x \in [0,40]$ as shown in Fig. \ref{fig:fit}. 

\section{Conclusion}
\label{sec:conclusion}

We described an approach to the Symbolic Regression problem combining a MAP-Elites exploration scheme together with a evolutionary optimizer. The optimizer, namely CMA-ES, allowed us to find ephemeral constants involved in symbolic expression with high accuracy. Starting from the same primitive set, we demonstrated high success rates on a large sample of reference targets taken from the literature.

We use the MAP-Elites~\cite{DBLP:journals/corr/MouretC15} algorithm to search for equations that are both accurate and diverse, across a range of selected topological features (parsimony, number of free scalars, number of non-algebraic functions). Additionally, other features could be taken into account to increase the diversity of the explored equations, catering either to equation topology (\eg number of trigonometric functions) or to the mathematical properties of the target function (\eg number of modes of the optimized equation compared to the target function, error measures on the first-order derivative, etc). An associated difficulty is the increase of number of bins in the MAP-Elites grid when more features are taken into account. This can make the algorithm focus too much on diversity which would reduce convergence speed. This can be prevented by using the CVT-MAP-Elites algorithm~\cite{vassiliades2017using}.

The use of CMA-ES as ephemeral constants optimizer is a computationally intensive technique. Thus, it may difficult to scale our methodology to higher-dimensional or more complex target functions, as CMA-ES would require larger evaluations budgets.
To overcome these limitations, alternative optimization techniques will have a key role to play, such as CMA-ES with several populations~\cite{hansen2009benchmarking} or variants of CMA-ES capable of handling large number of dimensions~\cite{loshchilov2018large}.
Moreover, to put the algorithm in practice on noisy targets (see \cite{miranda2017noisy}), further work will need to be done on the performance of black-box optimizers facing noise. In particular, because of its accuracy, CMA-ES is likely to return a large set of possible solutions that all seem to perfectly fit to the data. In this context, the handling of error bars throughout the entire SR process will be critical. Alternatively, we could employ optimization techniques that are inherently tolerant to noise, like Bayesian Optimization~\cite{frazier2018tutorial,pelikan2002scalability}.

The set of genetic operations was intentionally left in its most basic form. We wanted to explore how the combination of several methods can perform on SR problems. Our algorithms leaves a lot of room for improvement and we hope that state-of-the-art GP techniques together with refined analyses of feature selection will allow to achieve better performance.

With very few adjustments, the algorithm we described here can handle systems of partial derivative equations. We leave this aspect and the automated discovery of coupled differential equations in physical systems for future work.

\section*{Acknowledgments}
The work of Vincent Reverdy has been supported by the National Science Foundation under the grants SI2-SSE-1642411 and CCF-1647432. The work of Leo Cazenille has been supported by Grant-in-Aid for JSPS Fellows JP19F19722.

\bibliographystyle{unsrt}
\bibliography{mybiblio}

\end{document}